# Training large margin host-pathogen protein-protein interaction predictors

ABDUL HANNAN BASIT[2], WAJID ARSHAD ABBASI[1], AMINA ASIF[1], AND FAYYAZ UL AMIR AFSAR MINHAS[1, †]

[1]*Biomedical Informatics Research Laboratory, Department of Computer and Information Sciences, Pakistan Institute of Engineering and Applied Sciences (PIEAS), Nilore, Islamabad, Pakistan*
[2]*Department of Electrical Engineering, Pakistan Institute of Engineer-ing and Applied Sciences (PIEAS), Nilore, Islamabad, 44000, Pakistan.*
[hannan.ahb@gmail.com;](mailto:hannan.ahb@gmail.com) [wajidarshad@gmail.com](mailto:wajidarshad@gmail.com); [a.asif.shah01@gmail.com](mailto:a.asif.shah01@gmail.com); [†][afsar@pieas.edu.pk](mailto:afsar@pieas.edu.pk)

Detection of protein-protein interactions (PPIs) plays a vital role in molecular biology. Particularly, infections are caused by the interactions of host and pathogen proteins. It is important to identify host-pathogen interactions (HPIs) to discover new drugs to counter infectious diseases. Conventional wet lab PPI prediction techniques have limitations in terms of large scale application and budget. Hence, computational approaches are developed to predict PPIs. This study aims to develop large margin machine learning models to predict interspecies PPIs with a special interest in host-pathogen protein interactions (HPIs). Especially, we focus on seeking answers to three queries that arise while developing an HPI predictor. 1) How should we select negative samples? 2) What should be the size of negative samples as compared to the positive samples? 3) What type of margin violation penalty should be used to train the predictor? We compare two available methods for negative sampling. Moreover, we propose a new method of assigning weights to each training example in weighted SVM depending on the distance of the negative examples from the positive examples. We have also developed a web server for our HPI predictor called HoPItor (Host Pathogen Interaction predicTOR) that can predict interactions between human and viral proteins. This webserver can be accessed at the URL: http://faculty.pieas.edu.pk/fayyaz/software.html#HoPItor.

## 1. Introduction

Proteins are complex molecules that take part in virtually all life processes in living organisms.[1] They play a vital role in the processes carried out at the cellular level including, but not limited to, metabolism, decision making, and structural organization.[2] Protein sequences are composed of long chains of 20 amino acids.[1] The sequence of a protein determines its 3*D* structure, and, consequently, the structure determines its specific function.[1,3] Proteins rarely act alone; more than 80% of proteins operate in complexes.[4] Proteins perform different functions by interacting with one another [1,2,5]. Therefore, to understand the functionality of proteins, it is very important to find out about protein-protein interactions (PPIs).[6] A special type of PPIs is host—pathogen interactions (HPIs) that in which the pathogen imparts infectious diseases to its host.[7] According to the World Health Organization, each year more than 17 million people are killed by infectious diseases.[8,9] To fight these infectious diseases, it is important to identify HPIs as it is a key step in drug design.[6]

† Corresponding author.

Conventional wet lab techniques are expensive and time-consuming, making it almost impossible to assess all possible combinations of HPIs.[10,11] Therefore, computational approaches are used to predict HPIs.[10,11] For example, if we want to find possible interactions of only 2000 host proteins with 500 pathogen proteins, the possible host-pathogen combinations turn out to be one million. For this reason, there is a shortage of experimentally verified HPI data. Hence, computational approaches are valuable to predict putative HPIs.[11] Computational studies are, therefore, vital to increase the available PPI data and eventually increase the pace of research towards drug design.[11]

Most of the available computational methods to predict HPIs use sequence and structural similarity based techniques.[7,11,12,13,14] However, due to limited availability of structural information[15], sequence-based methods are better. Machine learning techniques are well known in bioinformatics to predict PPIs, and they use protein sequences to make feature vectors and then the model learns from available data to predict unknown interactions.[11,12,13,16,17] A binary classifier uses one host protein and one pathogen protein as a paired example.[18] The positive samples in the training of the classifier are experimentally verified whereas, the negative samples are generated computationally [11,13,18,19]. Support vector machines (SVMs) are being widely used as the machine learning model to predict PPIs [13,16,17] and are also adopted for this study.

The first question that arises is how to generate negative samples to get a good predictor. Ben-Hur *et al.*[19] discuss the available methods to choose negative examples and conclude that to avoid the bias in performance estimation of the classifier, negative samples should be selected uniformly at random. Although the random sampling may contaminate the negative examples with interacting proteins, this contamination is likely to be small. Whereas, a recent method called DeNovo negative sampling is proposed by Eid *et al.*[13] to replace random sampling. DeNovo sampling is a dissimilarity based negative sampling criterion that takes into account the sequence similarities of viral proteins that interact with a host protein $h$ before pairing host protein $h$ as a negative sample with a viral protein $v$ [13]. This study compares these two methods to see which one is better.

The second question is what should be the negative sample size. Should the number of the negative samples, $N$, be less than, equal to or greater than the positive samples, $P$? Eid *et al.* [13] and others [16,17] used $P:N = 1:1$ whereas, we propose that the entire data set ($N >> P$) should be used for training the classifier. In this study, we compare both strategies to find out the better one.

The third question arises that what type of margin violation penalty for the large margin machine learning should be used. Eid *et al.* [13] and others [16,17,18] use an un-weighted margin violation penalty for each class. The idea of using a weighted margin violation penalty for SVMs is widely implemented in other fields of science [20,21,22]. Ravikant *et al.*[23] used the idea of weighted penalties in an energy-based docking method for PPIs. We propose to use a weighted SVM in developing an HPI predictor. This study compares the two methods to see their effect on the learning ability of the SVM model.

Table 1. Partitioned Viral Families

| Group No. | Family Name | Positive Sample Size | Negative Sample Size |
|---|---|---|---|
| 1 | Paramyxoviridae | 762 | 1797 |
| 2 | Filoviridae | 114 | 592 |
| 3 | Bunyaviridae | 159 | 508 |
| 4 | Flaviviridae | 291 | 25953 |
| 5 | Adenoviridae | 88 | 3453 |
| 6 | Orthomyxoviridae | 664 | 5004 |
| 7 | Chordopoxviridae | 194 | 4158 |
| 8 | Papillomaviridae | 245 | 6665 |
| 9 | Herpesviridae | 1001 | 25505 |
| 10 | Retroviridae | 1399 | 8062 |
| | **Total** | **4917** | **81697** |

## 2. Methods

### 2.1. *Datasets and preprocessing*

HPIs are obtained from supplementary data provided by Eid *et al.*[13] This dataset is originally obtained from VirusMentha.[24] After removing duplicates, it has 4971 unique interactions between 2237 human proteins and 337 viral proteins. To simulate the novel environment, the viral proteins are divided into ten groups based on their biological families as enlisted in Table 1. The partitioned data was obtained from Eid *et al.*[13] on request.

### 2.2. *Generating Negative Samples*

Machine learning based PPI predictors require both positive and negative data sets for their training. As the available data of interactions is of the positive class only, the generation of negative examples is the first and foremost step. We evaluate two different techniques of generating negative samples in order to select the better one.

#### *2.1.1  Random Negative Sampling*

Ben-Hur *et al.*[19] argued that even though the random sampling may contaminate the data set, this contamination affects the model very little, and, thus, this sampling should be used. This method is illustrated in Fig. 1. First, a random host protein and a random pathogen protein are chosen; then, it is checked whether there exists a positive sample between them; if not, then the randomly chosen samples are selected as negative examples. The solid connectors represent the positive HPIs and the dotted connectors show the randomly selected negative HPIs. Pathogen protein 1 can be paired with host proteins *a* and *c* as a negative example; however, it cannot be paired negatively with host protein *b* since there exists a positive sample between 1 and *b*.

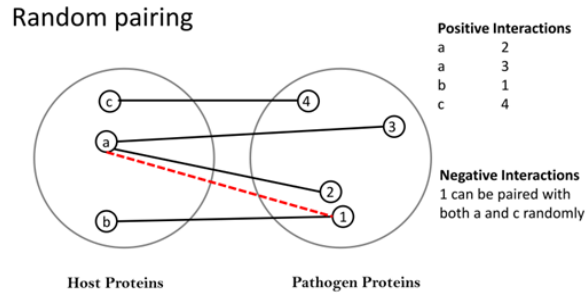

**Fig. 1.** An illustration of Random negative sampling

*2.1.2    DeNovo Negative Sampling*

Eid *et al.*[13] hypothesized that viral proteins having high similarity in their sequences can interact with a lot of similar host proteins. Based on this hypothesis, the authors argued that random negative sampling will result in a large number of false negative samples. To reduce false negatives, Eid *et al.*[13] presented a dissimilarity based negative sampling criterion called DeNovo negative sampling.

In DeNovo sampling, the sequence similarity of two viral proteins that interact with host proteins is first determined. If they are more similar than a cut-off value, a host protein that has a positive sample with one of these two viral proteins cannot be paired as a negative sample with the second one. Dissimilarity scores are used to compare the similarities between viral proteins, which are obtained by taking the complement of normalized bit scores. These bit scores are obtained by doing the all-vs-all global alignment of viral proteins. At a dissimilarity threshold $T$, the negative samples that do not fulfill the criterion are filtered out, and random sampling is done over the rest of the negative examples. We use $T = 0.7$ in this study. This method is illustrated in Fig. 2. The complete details of DeNovo pairing technique is given in.[13]

In Fig. 2, pathogen protein 1 cannot be paired with host protein b because it has a positive pairing with it. Moreover, it also cannot be paired with host protein $a$ as 1 and 2 are similar to each other, i.e., their dissimilarity distance is less than $T$, and 2 is pairing positively with $a$. However, viral proteins 1 and 4 have dissimilarity distance $\geq T$, and they have positive example with human proteins b and c respectively, therefore, 1 can only be paired with $c$ as a negative example.

## 2.3  Feature Extraction
The selected features are the same as incorporated by Eid *et al.*[13], originally proposed by Shen *et al.*[16] and others.[12, 17] The feature extraction is divided into two steps.

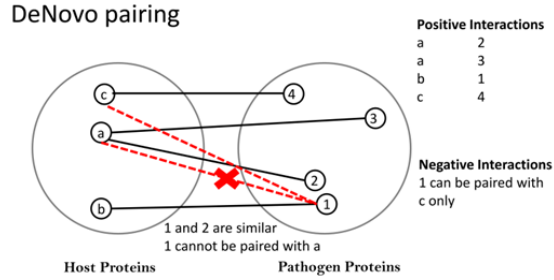

Fig. 2. An illustration of DeNovo negative sampling

### 2.3.1 Clustering

The 20 amino acids are first clustered into seven groups based on their physiochemical properties that affect the protein interactions most. These properties are volumes of side chains and dipoles. All 20 residues in protein sequences are replaced by their corresponding cluster number, thus, this process gives mapped protein sequences. The seven clusters are $\{A, V, G\}, \{I, L, F, P\}, \{Y, M, T, S\}, \{H, N, Q, W\}, \{R, K\}, \{D, E\}$ and $\{C\}$. For example, let's assume we have a protein sequence $APRIRGQ$. It will be mapped as $1252514$ because $A$ and $G$ belong to cluster 1, $P$ and $I$ belong to cluster 2, $R$ belongs to cluster 5, and, $Q$ belongs to cluster 4.

### 2.3.2 Frequency of 3-mers

After clustering, the frequency of all possible 3-mers is calculated in each protein sequence giving rise to a feature vector. The obtained feature vector is normalized to the range of $\{0,1\}$ for each protein independently. The two normalized feature vectors, one belonging to the host protein and the other belonging to the pathogen protein, of a positive or a negative class, are concatenated into a single feature vector to represent the interaction. The length of each protein's feature vector for 3-mers is $7^3 = 343$. Thus, the concatenated feature vector has length $343 + 343 = 686$.

## 2.4 Classification Model

Support Vector Machines (SVMs)[25] are used with radial basis function kernel in this study. SVM is the discriminant function that maximizes the geometric margin, $1/\|w\|$ or, equally, minimizes $\|w\|^2$. The optimization problem for soft SVM can be written as:

$$\min_{\mathbf{w},b} \frac{1}{2}\|\mathbf{w}\|^2 + C \sum_{i=1}^{n} \xi_i, \quad (1)$$

such that:

$$y_i(\mathbf{w}^T \mathbf{x_i} + \mathbf{b}) \geq 1 - \xi_i, \text{where } \xi_i \geq 0. \quad (2)$$

Here, $C$ is the margin violation penalty and it determines the relative significance of margin violation of all training examples and maximization of the margin. Moreover, $\xi_i$ is the margin error of $i$-th example allowing the example to be misclassified ( $\xi_i > 1$ ) and $y_i$ is the label of $i$-th example. The above equation can be expressed as the dual formulation:

$$\max_{\alpha} \sum_{i=1}^{n} \alpha_i - \frac{1}{2}\sum_{i=1}^{n}\sum_{j=1}^{n} y_i y_j \alpha_i \alpha_j \mathbf{x}_i^T \mathbf{x}_j, \quad (3)$$

such that:

$$\sum_{i=1}^{n} y_i \alpha_i = 0, \text{where } 0 \leq \alpha_i \leq C. \quad (4)$$

In (3), the term $\mathbf{x}_i^T \mathbf{x}_j$ is replaced by a kernel function, $K(i,j)$. In this study, radial basis function is used which is given as:

$$K(i,j) = \exp(-\gamma \times |i - j|). \quad (5)$$

### 2.5 Choosing Margin Violation Penalty

The choice of $C$ affects the SVM model as given in (3) and (4). Now, we will test two approaches to assign value of $C$.

#### 2.5.1    Un-weighted Margin Violation Penalty

In this approach[25], a constant value of $C$ is assigned to each class proportional to the size of that class. For example, if we choose $C = 1$, then $C_+ = 1/P$ and $C_- = 1/N$ where $P$ is number of positive training examples and $N$ is number of negative training examples. This means that the class that appears more often gets lower value of $C$ than the other one. However, the value of $C$ within that class remains the same.

#### 2.5.2    Weighted Margin Violation Penalty

In this approach[20], a separate value of $C$ is assigned to each training example. We used this idea of using weighted margin violation penalty in our study. In the formulation of SVM, (4) is replaced by the following:

$$\sum_{i=1}^{n} y_i \alpha_i = 0, \text{ where } 0 \leq \alpha_i \leq C w_i. \quad (6)$$

Here, $w_i$ is the weight of $i$-th example, which is multiplied by the constant value $C$ to get the effective margin violation penalty. Assignment of weights is discussed in next section in detail.

### 2.6 Confidence in Negative Samples

We used the idea of weighted margin violation penalty to quantify our confidence in each negative sample. To do this, we assigned weights, $w_i$, to each training example depending on how likely it is that a positive/negative example is truly a positive/negative example in the protein world. As the positive training examples are experimentally verified, therefore, all the positive examples are assigned equal weights proportional to their prior class probability. As the negative training examples are computationally generated, therefore, all the negative examples are assigned weights between 0 and 1 depending upon their probability of truly being a negative example. The confidence of a negative example, $(h, v)$, being truly negative is large if there are no viral proteins that interact with host protein $h$ and are similar to viral protein $v$.

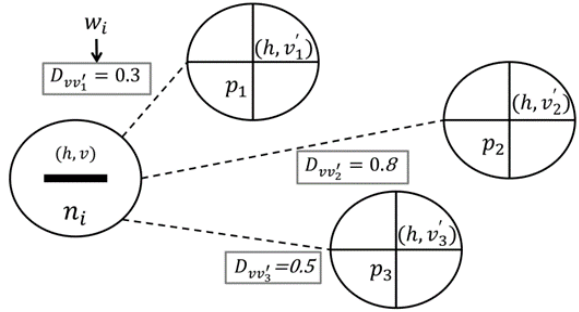

**Fig. 3.** An example of selecting margin violation penalty weight, $w_i$. $n_i$ is the $i$-th negative pair $(h, v)$ between host protein $h$ and viral protein $v$. Whereas, host protein $h$ has positive pairs with $v'_1, v'_2,$ and $v'_3$. Similarity of $v$ is maximum with $v'_1$ as compared to $v'_2,$ and $v'_3$. Therefore, dissimilarity distance $D_{vv'_1} = 0.3$ is set as $w_i$.

Therefore, we set $w_i$ equal to the dissimilarity score of the viral protein $v$ in negative training example $(h, v)$ to the most similar viral protein $v'$ that has a positive example $(h, v')$ with the host protein $h$. This idea is illustrated in Fig. 3. Weight of $i$-th negative example $n_i$ which is paired as $(h, v)$, is calculated by checking the dissimilarity distance $D$ of viral protein $v$ with the all other viral proteins $v'$ having a positive example $p$ with host protein $h$ and then selecting the least dissimilarity distance $D_{vv'}$ as $w_i$.

## 2.7 Performance Metrics

The performance metrics used in this study are the area under the Receiver Operating Characteristic curve (AUC-ROC) and area under the precision-recall curve (AUC-PR), as explained in [11] and [26]. True positive rate (TPR), also known as recall, is given as:

$$TPR = \frac{TP}{TP + FN},$$

whereas, false positive rate (FPR) is given as

$$FPR = \frac{FP}{TP + FN},$$

and precision is given as

$$Precision = \frac{TP}{TP + FP} .$$

AUC-ROC is the area under the plot between TPR and FPR at various thresholds. ROC curve tells us how good a predictor can detect true positives at a given rate of false positives. Moreover, AUC-PR is the area under the precision-recall curve (PR curve). The PR curve is sensitive to false positives while the ROC curve is not. The evaluation of the classifier is done using leave-one-group-out cross-validation as implemented in [13]. This method of cross-validation models the realistic scenario as no training data is available for a virus whose proteins are given in testing. Training parameters of the SVM (C and RBF kernel spread) were selected using nested 2-fold cross-validation.

## 3    Results and Discussion

### 3.1 Choosing Negative Sampling

We set-up two SVM models using Scikit-learn 0.17[27] in Python 2.7[28] to decide which sampling criterion is better. The first model is trained with negative examples generated by using Random negative sampling, while, the second model is trained with negative examples generated by using DeNovo negative sampling at $T = 0.7$. Eid *et al.* [13] used the same number of positive and negative examples, i.e. $P: N = 1: 1$. However, to simulate the real word scenario where $N \gg P$, we selected entire data set of DeNovo negative examples instead of using the reduced data set. The group-wise number of positive and negative examples is given in Table 1.

Table 2. Comparison OF AUC-ROC and AUC-PR at $C = 10$, $\gamma = 0.1$, and For DeNovo, $T = 0.7$

| Group No. | Random Negative Sampling | | DeNovo Sampling with Un-weighted $C$ | | DeNovo Sampling with Weighted $Cw_i$ | |
|---|---|---|---|---|---|---|
| | AUC-ROC | AUC-PR | AUC-ROC | AUC-PR | AUC-ROC | AUC-PR |
| 1 | 0.43 | 0.244 | 0.96 | 0.95 | 0.994 | 0.992 |
| 2 | 0.272 | 0.101 | 0.984 | 0.973 | 0.999 | 0.998 |
| 3 | 0.38 | 0.203 | 0.952 | 0.943 | 0.97 | 0.964 |
| 4 | 0.53 | 0.015 | 0.277 | 0.011 | 0.269 | 0.01 |
| 5 | 0.487 | 0.027 | 0.989 | 0.917 | 0.995 | 0.957 |
| 6 | 0.405 | 0.09 | 0.877 | 0.736 | 0.964 | 0.908 |
| 7 | 0.412 | 0.03 | 0.962 | 0.778 | 0.986 | 0.932 |
| 8 | 0.437 | 0.023 | 0.932 | 0.661 | 0.973 | 0.721 |
| 9 | 0.478 | 0.034 | 0.816 | 0.358 | 0.91 | 0.57 |
| 10 | 0.353 | 0.106 | 0.773 | 0.517 | 0.896 | 0.667 |
| Weighted Average | **0.47** | **0.05** | **0.68** | **0.39** | **0.73** | **0.50** |

In order to compare both SVM models, the same number of random negative examples is chosen for each group as that of DeNovo negative examples. The testing is done using 10-fold leave-one-group-out cross validation. The test examples for both models are kept the same for comparison purposes, i.e., the test examples are the ones generated using DeNovo negative sampling. The weighted average AUC-ROC for the model trained with random negative sampling is 0.47, while the weighted average AUC-ROC for the model trained with DeNovo negative sampling is 0.68. Moreover, the average AUC-PR for random sampling is 0.05, while it is 0.39 for DeNovo sampling. The detailed group-wise results are shown in Table 2. These significantly improved results show that *DeNovo sampling* is indeed better than *Random sampling*. Therefore, we choose DeNovo negative sampling for building our HPI predictor.

*3.2 Margin Violation Penalty*

After choosing the negative sampling method, we compare un-weighted margin violation penalty, i.e., constant each class and weighted margin violation penalty $Cw_i$ for each example. The data set and the rest of the parameters are the same as used for the DeNovo classifier in Section 3.1. The weighted average of AUC-ROC for un-weighted $C$ is 0.68 while it turns out to be 0.73 for weighted $Cw_i$. Group-wise improvement in AUC-ROC of all groups except group 4 can be seen in Table 2. There is considerable improvement in the AUC-ROC of last two groups. For group 9, it improved from 0.82 to 0.91, whereas, for group 10, it improved from 0.77 to 0.9.

**Table 3.** Comparison of SVM models trained with reduced data and full data

| Group No. | $N$ | Train with Reduced Data AUC-ROC | Train with Full Data AUC-ROC |
|---|---|---|---|
| 1 | 416 | 0.953 | 0.993 |
| 2 | 114 | 0.987 | 0.999 |
| 3 | 159 | 0.956 | 0.97 |
| 4 | 291 | 0.254 | 0.244 |
| 5 | 88 | 0.994 | 0.996 |
| 6 | 663 | 0.969 | 0.976 |
| 7 | 194 | 0.982 | 0.992 |
| 8 | 245 | 0.95 | 0.964 |
| 9 | 991 | 0.918 | 0.921 |
| 10 | 1087 | 0.823 | 0.9 |
| Total | **4248** | Wt-Avg: **0.87** | **0.9** |

The average AUC-PR for un-weighted $C$ is 0.39, while it is 0.5 for weighted ones. Improvement in AUC-PR of all groups except group 4 can also be seen in Table 2. There is considerable improvement in the AUC-PR of the last five groups. For example, AUC-PR of group 9 improved from 0.36 to 0.57, whereas, AUC of group 6 and group 7 improved from 0.74 to 0.91 and 0.78 to 0.93 respectively.

It is important to note that results of group 4 are consistently bad throughout the study. We investigated the reasons for this bad performance and found that the viral proteins in this group, belonging to Flaviviridae, are very dissimilar to the rest of the protein families. Therefore, the classifiers are unable to predict the quite different behavior of this group.

These results show that using weighted margin violation penalty is better, and, therefore, we selected this method to build our HPI predictor.

### 3.3 Test on Reduced Data

We also test the effect of reduced data on the SVM models. Here, we compare two SVM models trained with DeNovo sampling and weighted margin violation penalty. However, the first model is trained on the reduced data set, i.e., $P = 4971$ (same as previous) and $N = 4248$ (reduced from $81697$ ), whereas, the second model is trained with full data (same as in Section 2.4).

For comparison purposes, both the models are tested with the reduced data set using leave-one-group-out cross-validation on 10 groups. The results are shown in Table 3. The average AUC-ROC for the SVM trained and with reduced data is 0.87, whereas, it is 0.9 for the SVM trained with full data and tested with reduced data. This improvement shows that choosing entire data set ($N >> P$) for training the model is better than choosing reduced data set ($N \leq P$).

### 4. Web Server

We trained our final HPI predictor on the full data set using DeNovo negative sampling and weighted margin violation penalty. We made a web server of our HPI predictor for the biologists to check whether a human protein interacts with a viral protein or not. It is named HOst Pathogen Interaction Predictor (HOPITOR). It can be accessed through http://faculty.pieas.edu.pk/fayyaz/software.html#HoPItor for free.

### 5    Conclusions

In this paper, we tried to answer three questions that arise while developing a host-pathogen protein-protein interaction predictor. To summarize:
1. Choosing the entire data set (N>>P) is better than choosing the reduced data set (N≤P).
2. Using DeNovo negative sampling is better than using random negative sampling to generate negative examples.
3. Using weighted margin violation penalty is better than un-weighted margin violation penalty in the training of the model.

After reaching these conclusions, we developed a web server for our HPI predictor.

**Acknowledgments**

The authors thank Fatma-Elzahraa Eid, Virginia Tech, for providing all the relevant data for this study and her continuous support throughout this study. Wajid A. Abbasi is supported by a grant (PIN: 213-58990-2PS2-046) under indigenous 5000 Ph.D. fellowship scheme from the Higher Education Commission (HEC) of Pakistan.